\documentclass[a4paper]{article}
\usepackage{spconf,amsmath,graphicx}
\usepackage{amsmath,graphicx,bm, setspace}

\title{
A Re-ranker Scheme for Integrating Large Scale NLU models
}
\name{Chengwei Su, Rahul Gupta, Shankar Ananthakrishnan, Spyros Matsoukas}
\address{Amazon.com , USA}
\begin{document}
\ninept

\maketitle
\begin{abstract}
Large scale Natural Language Understanding (NLU) systems are typically trained on large quantities of data, requiring a fast and scalable training strategy.
A typical design for NLU systems consists of domain-level NLU modules (domain classifier, intent classifier and named entity recognizer). 
Hypotheses (NLU interpretations consisting of various intent+slot combinations) from these domain specific modules are typically aggregated with another downstream component. 
The re-ranker integrates outputs from domain-level recognizers, returning a scored list of cross domain hypotheses.
An ideal re-ranker will exhibit the following two properties: (a) it should prefer the most relevant hypothesis for the given input as the top hypothesis and, (b) the interpretation scores corresponding to each hypothesis produced by the re-ranker should be calibrated. 
Calibration allows the final NLU interpretation score to be comparable across domains. 
We propose a novel re-ranker strategy that addresses these aspects, while also maintaining domain specific modularity.
We design optimization loss functions for such a modularized re-ranker and present results on decreasing the top hypothesis error rate as well as maintaining the model calibration.
We also experiment with an extension involving training the domain specific re-rankers on datasets curated independently by each domain to allow further asynchronization. 
\end{abstract}

\noindent\textbf{Index Terms}: Re-ranking, calibration, multi-task learning 
\section{Introduction}
\label{sec:intro}

Voice-controlled smart agents \cite{sarikaya2017technology} provide powerful ways for humans to interact with the underlying intelligent systems in performing various kind of tasks, such as playing music, asking for local directions to more recently having extended multi-turn conversations \cite{farber2016amazon}. 
Natural Language Understanding (NLU) is a crucial part of this system that interprets the user request and produces structured representation of the user's intention and extract units of information in the request.
Design of such large scale NLU systems requires a fast and scalable training with capabilities such as asynchronous and parallel training \cite{suendermann2009rule, glass1999challenges}. 
One of the NLU design approaches is a system modularized into domain-specific components (namely, domain classifier, intent classifier and named entity recognizers), where each domain represents a core service such as Music and Q\&A. 
Given this modularized system, we aim to design a combination model that merges the outputs from the domain-specific components and returns a scored list of cross-domain NLU hypotheses. 
In this work, we propose an NLU re-ranker model for re-scoring domain hypotheses, while maintaining the domain-wise modularity.
A domain specific re-ranker can allow a parallelized as well as asynchronous training across domains. 
In such a case, scores for the generated NLU hypotheses should capture the notion of correctness, as well as being calibrated across each domain for a direct comparison of cross domain scores. 
With calibration, the highest scoring hypothesis amongst the list of all domains' hypotheses can directly be assigned as output of the NLU system.
Calibration also allows creation of a sorted list of cross domain hypotheses based on the hypotheses scores. 
We demonstrate the application of this design of re-ranker on an Alexa use case (Alexa is the virtual assistant designed for Amazon devices), with improvements in NLU accuracy over a baseline involving unweighted combination of outputs from the domain-specific components. 

Several previous works have investigated the integration of hypotheses from multiple domain components using a combination model.
For instance, Robichaud et al. \cite{robichaud2014hypotheses} present a hypotheses ranking system to combine outputs from similar modular domain specific models, based on Lambda Rank Gradient Boosted Decision Trees \cite{burges2007learning}.
The combination model is a single model trained on outputs from each group of domain models and consequently, is shared across all the domains.
Crook et al. \cite{crook2015multi} extend a similar model to a multi-lingual setting.
Re-ranking of hypothesis coming from a system has been approached in other problems such as machine translation \cite{shen2004discriminative}, obtaining correct NLU hypotheses given multiple speech recognition hypotheses \cite{morbini2012reranking}, as well as re-ranking speech recognition hypotheses themselves \cite{sak2011discriminative}. 
On the other hand, researchers have also focused on designing re-ranking algorithms such as LambdaMART \cite{burges2007learning}, Adarank \cite{xu2007adarank} and Mcrank \cite{li2008mcrank}.
A few other noteworthy approaches that minimize error metrics based on a re-ranker approach inlude corpus weight estimation \cite{matsoukas2009discriminative}, minimum-risk training on translation forests \cite{li2009first}, batch tuning for statistical machine translation \cite{cherry2012batch} and minimium risk annealing \cite{smith2006minimum}.
In particular, the Yahoo! learning to rank challenge \cite{chapelle2011yahoo} led to several advances in hypothesis ranking.
Despite providing promising results on multiple tasks, the methods propose training a single model as a combination strategy, which will not maintain modularity of the system.
Also, the score yielded by the above methods is not calibrated for comparison across domains. 
Calibration \cite{bella2010calibration} is an important desired property of machine learning systems, in particular when multiple models are operating in conjunction \cite{zhong2013accurate}.
As we propose training modularized re-rankers, calibration becomes a critical requirement for selecting top hypotheses across all the domains. 
Calibration is also desirable by downstream components (for instance, for making decisions such as to accept or reject the top hypothesis). 
The novelty of our work lies in achieving modularization with the aid of calibration. 
The proposed design provides advantages such as asynchronous training across each domain, needing inputs from only the corresponding domain components.

NLU systems modularized as per the concept of domains have been proposed in several previous works \cite{sarikaya2016overview,robichaud2014hypotheses}. 
Domain Classifiers (DC), Intent Classifiers (IC) and Named Entity Recognizers (NER) are statistical models that outputs labels (in their corresponding label spaces) and corresponding confidence scores.
Hypothesis for a given domain is an aggregation of outputs from the corresponding domain-specific DC, IC and NER models. 
We design loss functions to jointly achieve the top hypothesis correctness along with calibration.
We also present analysis on the hypotheses calibration and perform an additional experiment involving training the domain specific re-rankers on datasets generated independently for each domain.
This experiment explores further asynchronization across domains where each domain is free to curate their own training data (as opposed on relying on a common data source before kicking off model training). 

This paper is organized as follows: In the next section, we provide a background of a typical design for DC, IC and NER models, followed by a description of the proposed re-ranker model.
Section 3 describes the re-ranker design, followed by experiments and results in section 4 and 5, respectively.
We discuss the additional experiment on re-ranker training using different datasets in section 6 and present the conclusion in section 7. 


\section{Background: NLU system design}
\begin{figure}
\includegraphics[scale=0.21]{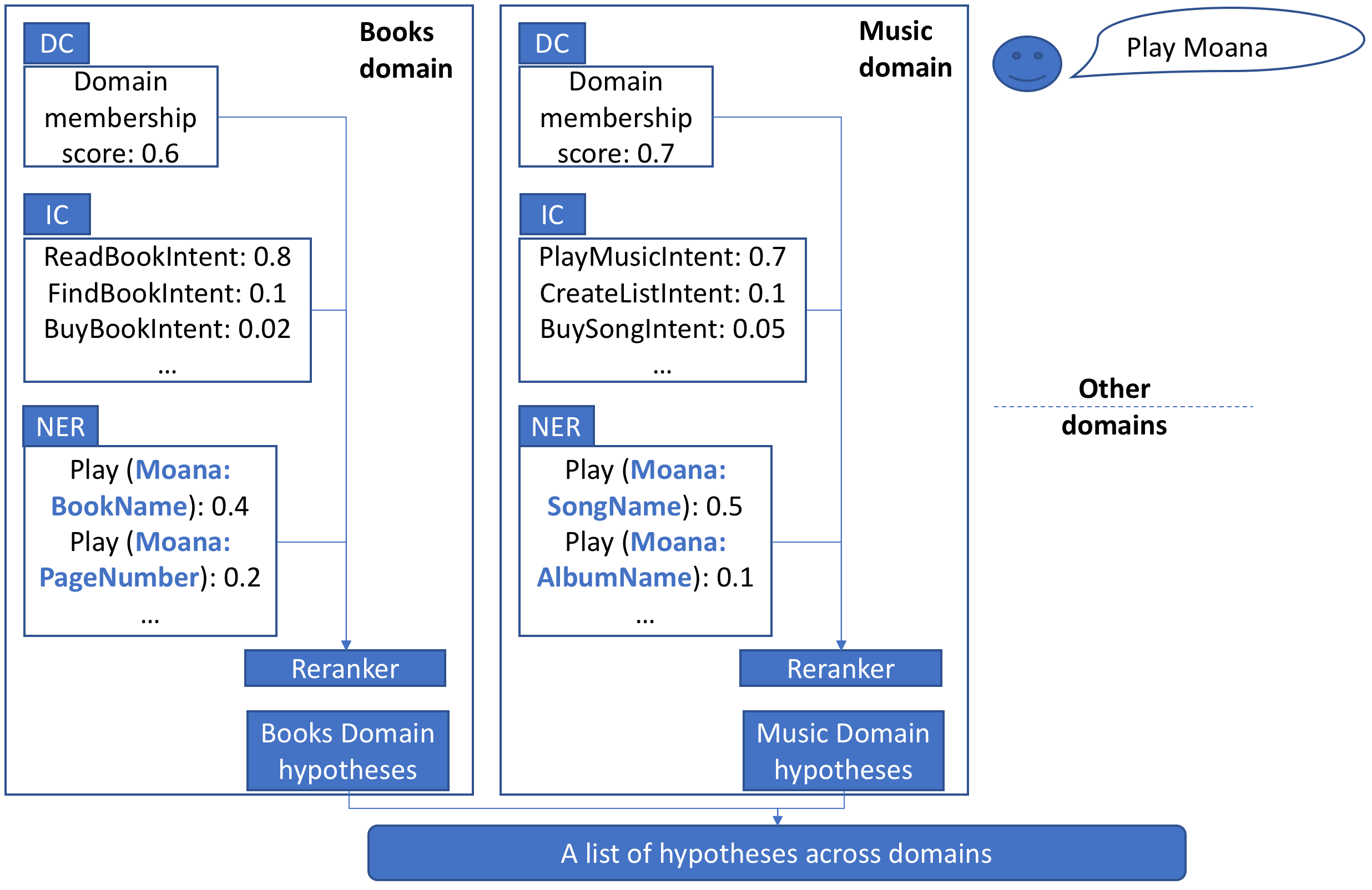}
\vspace{-3mm}
\caption{A description of a modularized NLU system design containing DC (One Vs All MaxEnt), IC (Multi class MaxEnt) and NER (CRF model) models. We propose the re-ranker as the combination scheme to obtain a ranked list of NLU hypotheses.}
\label{fig:example_figure}
\end{figure}

An effective NLU design approach involves modularization into domain components, with each component containing a set of three statistical models: (i) a Domain Classifier (DC), (ii) an Intent Classifier (IC) and, (iii) a Named Entity Recognizer (NER).
Below, we describe the NLU system used for our experiments and Figure~\ref{fig:example_figure} provides a pictorial description of the modularized system.

(1) Domain Classifier (DC): A domain classifier is a binary classifier indicating if a given utterance query is intended for the target domain. 
In our system, the DC is implemented, without loss of generality, as a Maximum Entropy (MaxEnt) classifier operating on n-grams extracted from the query utterance.
In Figure~\ref{fig:example_figure}, Books and Music domain DC yield scores of 0.6 and 0.7 for an incoming utterance ``play Moana".
A lower score is expected from an unrelated domain such as sports.

(2) Intent Classifier (IC): Each domain component serves multiple intents related to it.
For instance, a request relevant to the music domain component, could be about playing a specific song or adding a song to a playlist.
Therefore, each domain component should also be designed to return an intent specific score for a set of intents supported by that domain.
We train a multi-class MaxEnt model for intent classification, with the prediction targets set as the intents supported by that domain.
The model uses utterance n-grams as features.
In Figure~\ref{fig:example_figure}, books IC returns scores for associated intents such as ReadBookIntent and FindBookIntent, whereas Music IC returns scores for PlayMusicIntent and CreateListIntent.

(3) Name Entity Recognizer (NER): Finally, the entity recognizer serves the specific purpose or identifying named entities within a query.
For instance, for the query described in Figure~\ref{fig:example_figure}, ``Moana'' has to be recognized as album name if the query corresponds to the music domain, book name if the query corresponds to the books domain or a movie name if the query corresponds to the video domain.
We perform this task using a Conditional Random Field (CRF) classifier trained on utterance n-grams.
For each domain component, there is a domain specific set of labels on which the NER model is trained.

\section{Re-ranker design}
In order to maintain the modularity of re-ranker across domains, we design domain-specific re-rankers with input from the corresponding domain specific components.
Given outputs from DC, IC and NER from each domain's statistical models, we first create a list of domain specific candidate hypotheses.
In this section, we describe the strategy for creating this list of candidate domain hypotheses.
These hypotheses and the corresponding DC, IC and NER scores are then used as input to the domain's re-ranker model.

\subsection{Generation of domain specific hypotheses} 
\begin{figure}
\includegraphics[scale=0.21]{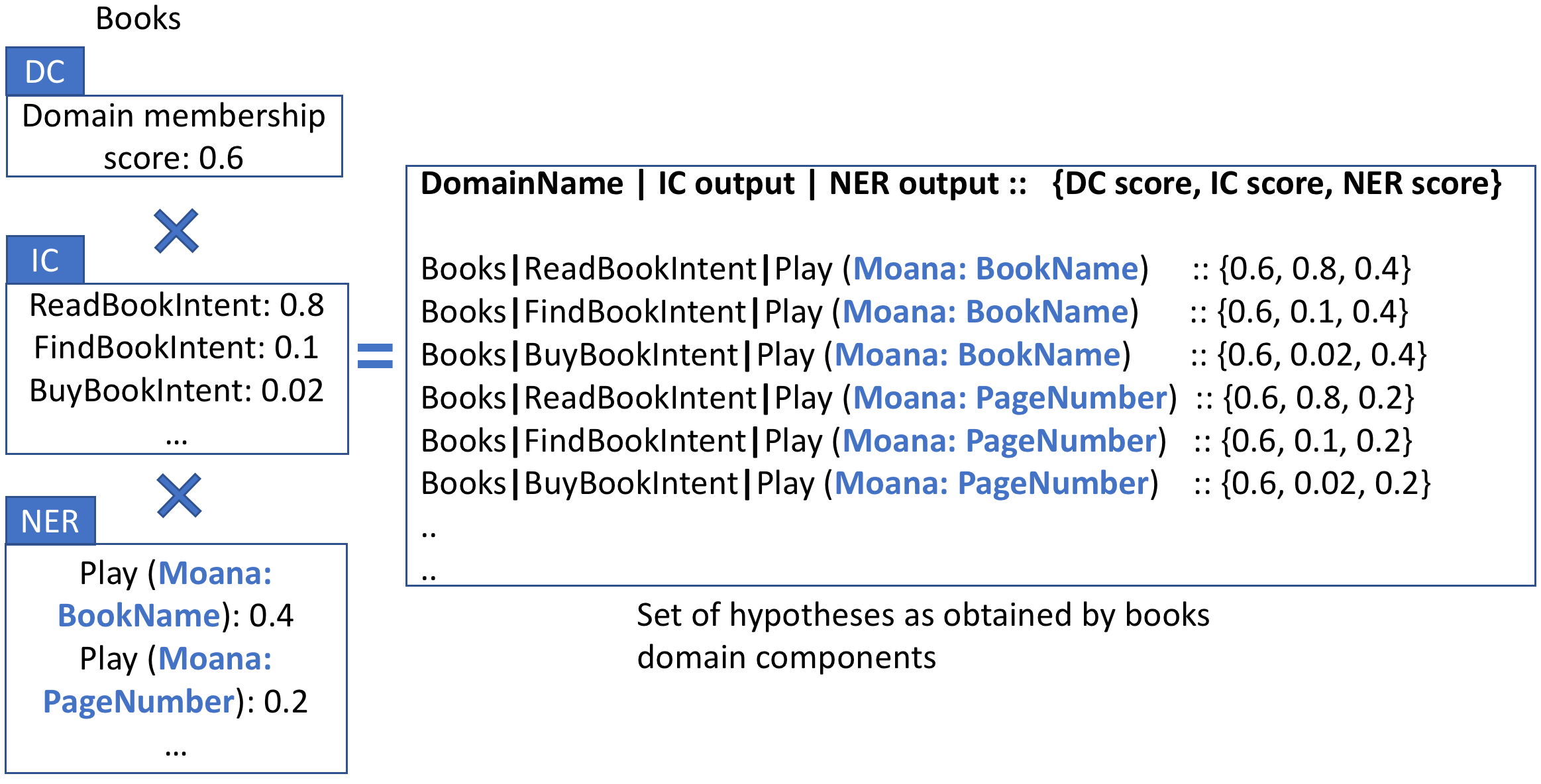}
\vspace{-7mm}
\caption{An example of set of hypotheses obtained from the Books domain components for the utterance ``play Moana". Note the scores corresponding the DC, IC and NER are retained for every generated hypothesis.}
\label{fig:hyps_figure}
\end{figure}

We initially create a list of in-domain hypotheses, which is obtained as a Cartesian product of the label outputs from DC, IC and NER models. 
An example of candidate domain hypotheses for Books domain is shown in Figure~\ref{fig:hyps_figure}.
A similar in-domain hypothesis generation process is repeated for all other domains.
Typically the Cartesian product is done using a beam search method, where only a limited set of most confident IC and NER hypotheses are considered.
Note that the scores from the DC, IC and NER for the corresponding labels are retained and are used as inputs to the re-ranker. 
Next, we define error metrics on a given hypothesis.

\subsection{Error definition on NLU hypotheses}
Given the set of domain hypotheses as described above, we define two errors metrics: (a) Semantic Error Rate (SemER) and, (b) Interpretation Error (IE).

\begin{figure}
\includegraphics[scale=0.25]{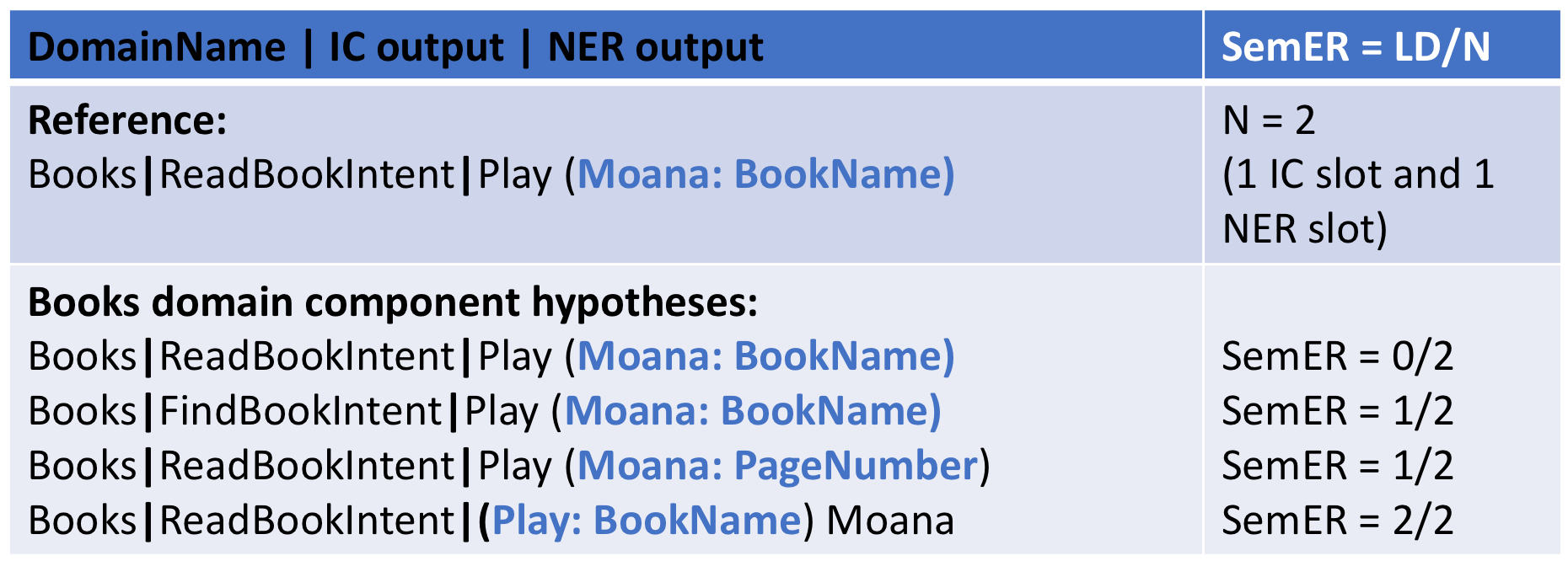}
\vspace{-3mm}
\caption{An example of SemER computation for hypotheses from the Books domain component. Note that the DC output is not used to compute Leveshtein distance (LD). N is the number of slots in the annotations.} 
\label{fig:ser_example}
\end{figure}

{\bf Semantic Error Rate} (SemER): Given the ground truth annotation for an utterance, we first compute the Levenshtein distance \cite{heeringa2004measuring} between the ground truth annotation and the domain hypothesis.
The slots used for computing the Levenshtein distance are the IC and NER labels.
We do not use DC output for distance computation as IC label space are non-overlapping across domains. 
Hence, an incorrect DC assignment automatically implies an incorrect IC assignment.
The computed distance is then normalized by the total slot count in the ground truth reference.
An example for sample hypotheses from the Books domain component is shown in Figure~\ref{fig:ser_example}.  
We represent the SemER value for a hypothesis $i$ corresponding to an utterance $u$ as SemER$_{ui}$.

{\bf Interpretation Error} (IE): IE is a hard metric that indicates if a hypothesis is correct or not, i.e., hypothesized intent and slots exactly match the ground truth. 
IE is 0 if SemER is 0 and 1 otherwise.
We represent the IE value for a hypothesis $i$ corresponding to an utterance $u$ as IE$_{ui}$.
Next, we present the design of our modularized re-ranker model as well as the optimization loss definitions based on the SemER and IE error metrics.

\subsection{Re-ranker model and Loss definition}
In order to maintain the modularity of the re-rankers per domain, we train a re-ranker for each domain.
During training, the domain's re-ranker training only receives hypotheses generated from the corresponding domain components on a re-ranker training dataset.
The training dataset, however, may contain ground truth annotations that may belong to other domains. 
Note that this implies that the in-domain hypotheses for utterances belonging to other domains will always be incorrect during the re-ranker training. 
After obtaining the set of domain hypotheses and the corresponding DC, IC and NER scores, we obtain the final hypotheses score as a function $f$ of the three scores with a parameter set $\bm w_{\mathsf d}$, as shown in equation~(\ref{eq:hyp_score}).

\begin{equation}\label{eq:hyp_score}
s^{ui}_{\mathsf d} = f(\bm w_{\mathsf d}, \bm l^{ui}_{ \mathsf d}) 
\end{equation} 

$s^{ui}_{\mathsf d}$ represents the score corresponding to the $i^\text{th}$ hypotheses for a domain $\mathsf d$, given an utterance $u$.
$\bm l^{ui}_{\mathsf d}$ is a vector representing the scores ($\log$ probability) returned by the set of domain specific statistical models (DC, IC and NER) for hypotheses $i$ for domain $\mathsf d$, given the utterance $u$. 
Note that we obtain a separate set of weights $\bm w_{\mathsf d}$ for each domain $\mathsf d$, thereby maintaining the modular nature of the NLU models, allowing parallel and asynchronous training. 

In order to obtain the weights $\bm w_{\mathsf d}$ for the weighted combination stated above, we experiment with cost functions that yields a high score $s^{ui}$ for domain hypotheses with low SemER while also bieng calibrated.
Based on the SemER and IE metrics, we define the following cost functions. 

{\bf Expected SemER loss (E-SemER)}: 
We define the E-SemER loss in equation~\ref{eq:joint_opt1}.
The expected SemER loss $S^{u}_{\mathsf d}$ encourages hypotheses with a lower SemER value to yield a higher score $s_{\mathsf d}^{ui}$ (and correspondingly higher $p_{ui}$).
The SemER objective quantifies the degree of hypothesis correctness and the objective attains minimum value for the utterance $u$ if $s_{\mathsf d}^{ui}$ values are in the reverse sorted order as the SemER$_{ui}$ values (hypothesis with the highest SemER$_{ui}$ obtains the least $s_{\mathsf d}^{ui}$).
Also, expected SemER is a better metric over alternates that focus on the correctness of the top hypothesis, as it provides a ranked list of hypotheses arranged by their degree of correctness. 

\begin{equation}\label{eq:joint_opt1}
\begin{aligned}
S^{u}_{\mathsf d} = \lambda_1 \sum_{ { \substack {{i \in {\text{Set of hypotheses}}} \\ \text{for $u$ in domain ${\mathsf d}$}}}} \underbrace{p^{ui}_{\mathsf d} \times {\text{SemER}_{ui}}}_{\substack{\text{Expected SemER loss} }}  \\ 
\end{aligned}
\end{equation}

\begin{equation}
\begin{aligned}
&\text{where;}\;p^{ui}_{\mathsf d} = \frac{\exp(s^{ui}_{\mathsf d})} { \sum_{ \substack {{i \in {\text{Set of hypotheses}}} \\ \text{for $u$ in domain ${\mathsf d}$}}} \exp(s^{ui}_{\mathsf d})} \\ 
&\lambda_1 = \frac{1}{\text{Utterances belonging to domain $\mathsf d$ in re-ranker training set}} \\ 
\end{aligned}
\end{equation}

{\bf Expected Cross entropy loss}: The objective with the cross-entropy loss $C^{u}_{\mathsf d}$ with respect to the labels IC$_{ui}$ is added for a calibration purpose.
Despite training the domain specific weights, the scores $s^{ui}_{\mathsf d}$ across the set of domains should be comparable.
Cross entropy loss allows machine learning model's outputs to be interpreted as Bayesian probabilities \cite{zadrozny2001obtaining}, allowing output scores from independently trained models to be directly comparable.
It also offers score refinement, by assigning a low score to an incorrect hypothesis and vice-versa. 
However, note that this loss does not take into account the granular SemER score into account for ranking, but only the coarse IE values.
Hence, all incorrect hypotheses are considered equally bad without regards to the degree of incorrectness.
Also, note that in equation~\ref{eq:joint_opt2}, we seek calibration for all the generated hypotheses and not just the top-best.

\begin{equation}\label{eq:joint_opt2}
\begin{aligned}
C^{u}_{\mathsf d} = \lambda_2 \hspace{-2mm} \sum_{ { \substack {{i \in {\text{Set of hypotheses}}} \\\text{for $u$ in domain ${\mathsf d}$}}}} \hspace{-3mm}  p^{ui}_{\mathsf d} \times \underbrace{((1-\text{IE}_{ui}) \log r^{ui}_{\mathsf d} + \text{IE}_{ui} \log (1-r^{ui}_{\mathsf d}))}_{\substack{{\text{Cross Entropy with IE$_{ui}$ as target}}}} 
\end{aligned}
\end{equation}

\begin{equation}
\begin{aligned}
&\text{where;}\; r^{ui}_{\mathsf d} = \sigma(s^{ui}_{\mathsf d}) \\
&\lambda_2 = \frac{1}{\text{Utterances in the re-ranker training set}} 
\end{aligned}
\end{equation}

$\sigma$ is the standard sigmoid function. 
Next, we define the various optimization schemes we use to learn the re-ranker weights $\bm w_{\mathsf d}$. 

\subsection{Re-ranker model optimization}
\label{sec:rmo}
We chose a linear function of the form $f(\bm w_{\mathsf d}, \bm l^{ui}_{ \mathsf d}) = \bm w_{\mathsf d}^T \bm l^{ui}_{ \mathsf d}$.
The reason for choosing a linear model is fact that the individual model components (DC, IC and NER) perform at high accuracy levels (DC models yield F-scores $>$ 95\% for each domain individually, IC models yield unweighted F-scores $> 90\%$ per domain and NER models yield F-score $>$ 88\% per named entity detection). 
Another off-line experiment using a shallow neural network instead of linear model does not show significant gains, reflecting that the DC, IC and NER scores are very well linearly correlated with SemER and IE values. 
A linear combination model has also been tested in several related re-ranking tasks such as machine translation \cite{matsoukas2009discriminative} and speech recognition \cite{chen1996empirical}. 
We experiment with four different re-ranker optimization models as described below.

\subsubsection{Baseline: Uniform re-ranker}
The first re-ranker model used in our experiments is a re-ranker that assigns uniform weights as $\bm w_{\mathsf d}$. 
The baseline strategy is the most naive model that retains domain specific modularization and does not require any additional training on top of the DC, IC and NER training. 
We also observe that this strategy performs competitively due to high accuracy of the DC, IC and NER models. 
The baseline strategy also approximates the following.

\begin{equation}\label{eq:prob}
\begin{aligned}
& \log P(\text{Hypothesis} | u) =  \log P(\text{Domain}, \text{Intent}, \text{Named-Entity} | u) \\ 
& \approx \log \underbrace{P(\text{Domain}|u)}_\text{DC score} +  \log \underbrace{P(\text{Intent}|u)}_\text{IC score} + \log \underbrace{P(\text{Named-Entity}|u)}_\text{NER score} \\
& = \bm{1^T} \bm l^{ui}_{ \mathsf d}
\end{aligned} 
\end{equation} 

\subsubsection{R1: Re-ranker trained using E-SemER loss only}
In the optimization R1, we only use the E-SemER loss for training as discussed in the previous section.
Note that this optimization is only run on in-domain data.
We produce domain specific hypotheses in training re-ranker for a given domain, hence hypotheses for ``out-of-domain" utterances are always incorrect. 
SemER$_{ui}$ in that case does not represent the quality of hypothesis and is merely a function of the count of NER labels. 
This optimization can be considered equivalent to other ranking based metrics in the literature that solely depend upon a ranking based loss (Lambda Rank Gradient Boosted Decision Trees \cite{burges2007learning}). 
The optimization is stated below.

\begin{equation}\label{eq:final_opt}
\bm w_{\mathsf d}^* = \arg\min_{\bm w_{\mathsf d}} \big( \hspace{-2mm}  \sum_{u \in {\substack{\text{Re-ranker training } \\ \text{ set in {$\mathsf d$}} }}} \hspace{-2mm} S^{u}_{\mathsf d} \big) 
\end{equation}

\subsubsection{R2: Re-ranker trained using E-CE loss only}
In the optimization R2, we use the E-CE loss computed on each utterance in the training set.
In computing E-CE loss, all incorrect hypotheses are labelled 0. Consequently, all hypotheses from ``out-of-domain" utterances are labelled incorrect, while only one hypothesis from in-domain utterances can be correct. 
We state the R2 optimization below. 

\begin{equation}\label{eq:final_opt}
\bm w_{\mathsf d}^* = \arg\min_{\bm w_{\mathsf d}} \big( \hspace{-2mm} \sum_{u \in {\text{Re-ranker training set}}} \hspace{-2mm} C^{u}_{\mathsf d} \big) 
\end{equation}

\subsubsection{R3: Re-ranker trained using E-SemER + E-CE loss}
In this scheme, we optimize a combination of both the losses as stated in equation~\ref{eq:final_opt}. 
We set $\lambda_1$ and $\lambda_2$ are set as show in equation~\ref{eq:lambda}, assigning equal importance to the expected SemER and the cross entropy losses. 

\begin{equation}\label{eq:final_opt}
\bm w_{\mathsf d}^* = \arg\min_{\bm w_{\mathsf d}} \big( k_1 \hspace{-3mm} \sum_{u \in {\substack{\text{Re-ranker training } \\ \text{ set in {$\mathsf d$}} }}} \hspace{-2mm} S^{u}_{\mathsf d} + k_2 \hspace{-2mm} \sum_{u \in {\text{Re-ranker training set}}} \hspace{-2mm} C^{u}_{\mathsf d} \big) 
\end{equation}

Where, $k_1$ and $k_2$ are hyperparameters tuned to account for the scaling differences between E-SemER and E-CE losses. 

\subsection{Obtaining the final hypothesis for a test utterance}
\label{sec:merge}
Given a set of domain specific DC, IC, NER and re-rankers, we decode a test utterance on a per-domain basis.
Each domain yields its set of domain specific hypotheses with corresponding hypotheses scores.
They are merged across all domains and sorted based on the interpretation scores to obtain the n-best for the utterance. 

We discuss our experimental setup in the next section, followed by results and discussions. 

\begin{figure}
\includegraphics[trim={1.4cm 8.5cm 0 3cm},clip,scale=0.4]{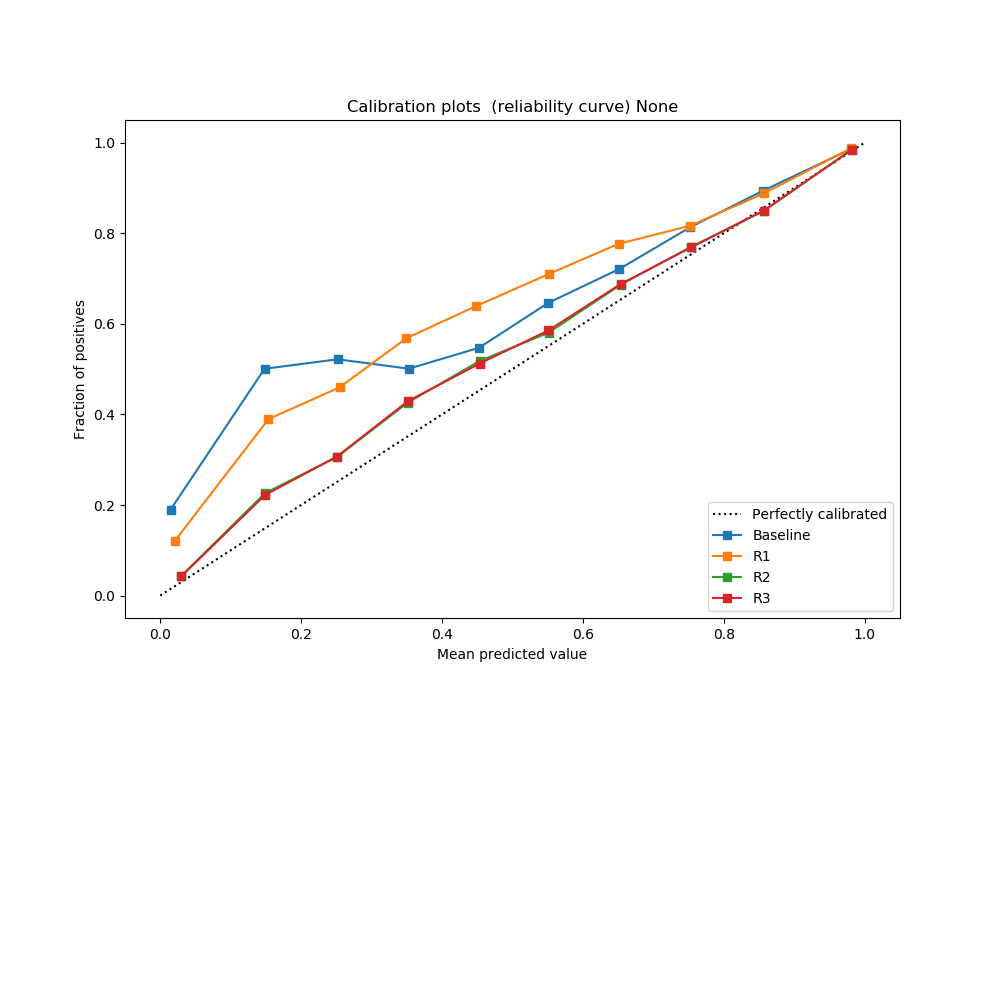}
\vspace{-7mm}
\caption{Confidence calibration plots obtained on hypotheses scores across all domains. While baseline and R1 optimization do not yield a good calibration, optimization using R2 and R3 provide the best calibration.}
\label{fig:calib_figure1}
\end{figure}

\begin{figure*}[t]
\hspace{-2mm}\includegraphics[trim={1.5cm 8.5cm 0 2.5cm},clip,scale=0.28]{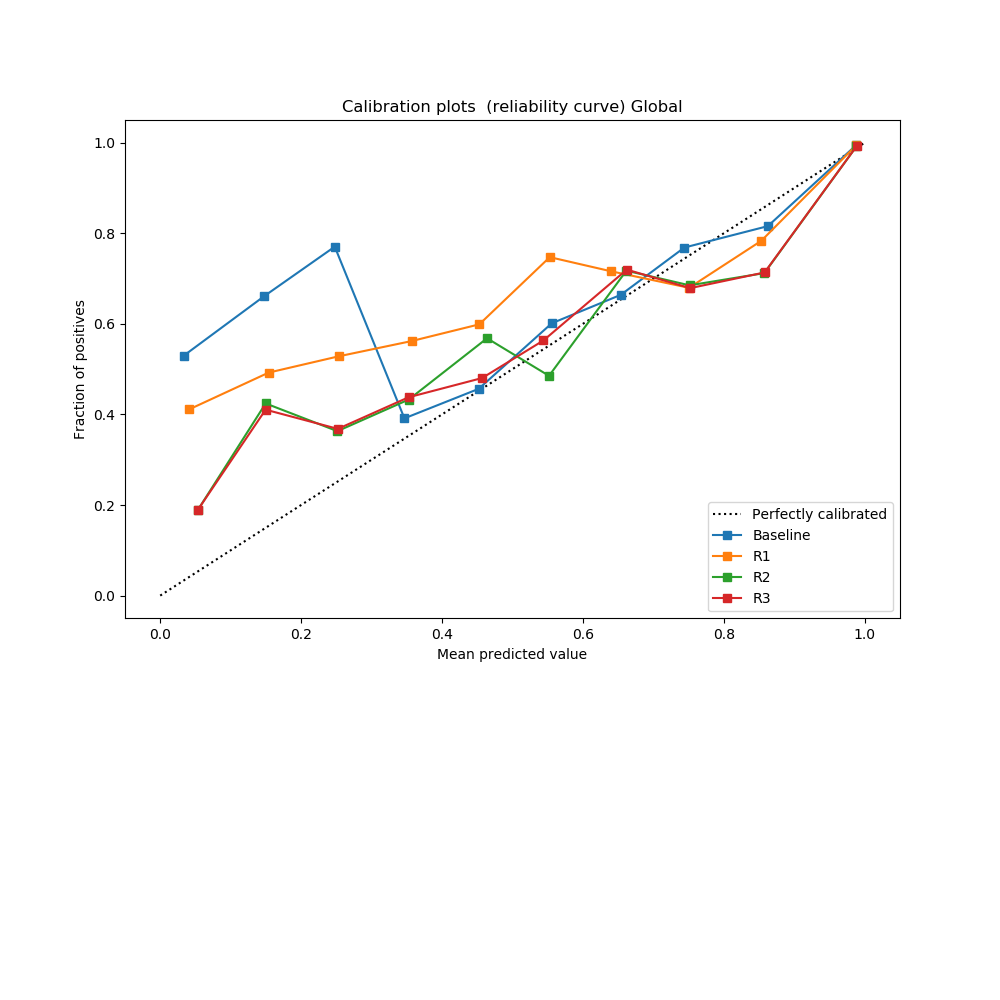} 
\hspace{-5mm}\includegraphics[trim={2.4cm 8.5cm 0 2.5cm},clip,scale=0.28]{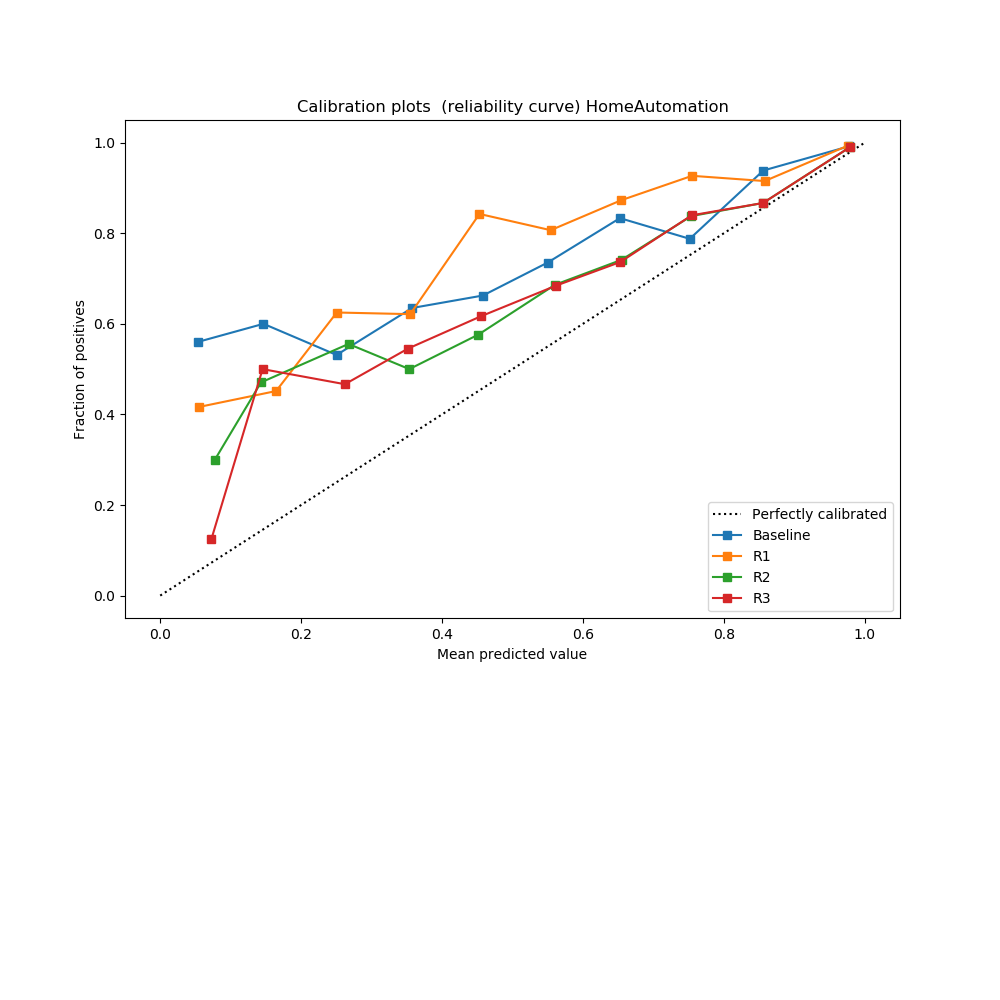} 
\hspace{-5mm}\includegraphics[trim={2.4cm 8.5cm 0 2.5cm},clip,scale=0.28]{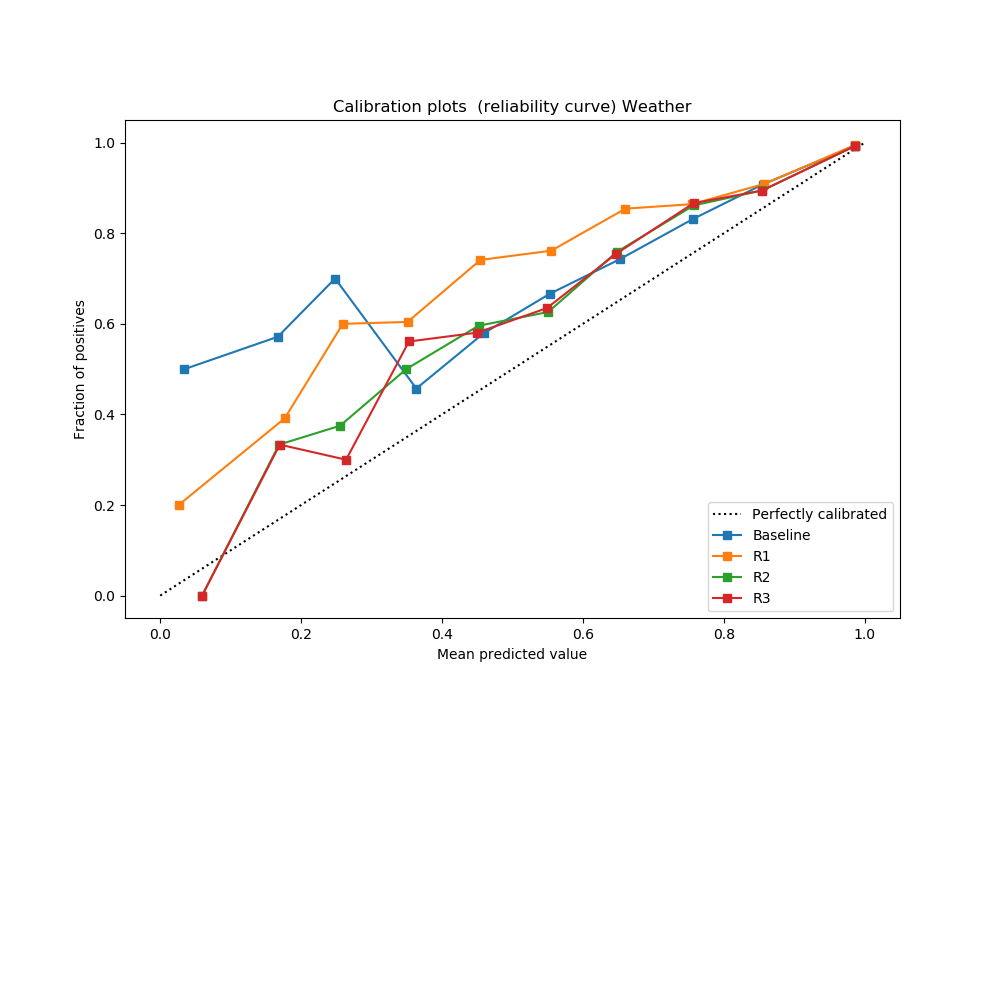} 
\vspace{-7mm}
\caption{Confidence calibration plots obtained on hypotheses on three domains: Global, HomeAutomation and Weather. The calibration varies per domain, particularly at lower thresholds.} 
\label{fig:calib_figure2}
\end{figure*}

\section{Experiments}
We test the various re-ranker training strategies on an NLU system serving an Alexa-enabled device.
We use $\sim$10M annotated utterances representative of user requests directed to this device.
We initially train the domain specific DC, IC and NER models on the training partition corresponding to the device. 
DC for each domain $\mathsf d$ is trained on the entire training partition with in-domain utterances labels as ``in domain $\mathsf d$'' and ``out of domain $\mathsf d$'' otherwise.
IC for a domain $\mathsf d$ is trained on the training data partition corresponding to that domain, with targets as the set of intent labels in that domain (``out of domain $\mathsf d$'' data for DC is not considered in IC training, as it does not contain intents corresponding to the domain $\mathsf d$).
Hyperparameters such as n-grams and regularization are tuned on the development partition. 
Similarly, NER for a domain $\mathsf d$ is trained on the domain specific training data partition, with target space spanning the set of named entities corresponding to that domain. 
After training the DC, IC and NER models, we train the domain-specific re-ranker on the same development partition using the optimizations stated in section~\ref{sec:rmo}.





\section{Results}
We present the baseline results as well as the relative improvements over the uniform weight baseline for the device in Table~\ref{tab:results1}.
Firstly, we observe that the baseline re-ranker model yields a low SemER value, indicating a strong performance. 
Over the strong baseline, we further obtain a significant improvement in the test performances (p-value $< 5\%$, bootstrap re-sampling test \cite{koehn2004statistical}) using each of the R1, R2 and R3 schemes.
R3 performs the best, suggesting a combination of a pure re-ranking based loss and calibration loss should be used.
However, our evaluation is based only on the correctness of the top best and the incorporation of the E-SemER loss may also be desirable from an n-best quality perspective.


\subsection{Calibration analysis}
Calibration is an important and desired property of machine learning models and is required in SLU systems from two aspects: (i) a well calibrated model observes less perturbations in performance with each model update and, (ii) several downstream components can use the score $s^{ui}_{\mathsf d}$ as true estimate of the model's confidence.
Moreover, in our case, calibration is required as hypotheses across domains are merged and sorted. 
We evaluate the confidence calibration of models with calibration plots: indicating the proportion of correct hypothesis in a bin and mean of scores in that bin.
We divide probability scores into 10 bins and compute mean of the scores $s^{ui}_{\mathsf d}$ for utterances in each of the bins.
The mean is plotted against the fraction of correct hypotheses (IE=0) in the corresponding bins. 
By definition, a perfectly calibrated curve is the one where mean of scores in the bin is equal to the proportion of correct hypothesis in that bin. 
For each re-ranker model (baseline, R1, R2 and R3), we present calibration curves in two different settings: 
(i) a cross-domain calibration curve: on top hypotheses obtained after aggregation from all domains (Figure~\ref{fig:calib_figure1}) and, 
(ii) per-domain calibration curves: on the subset of hypotheses obtained as top hypotheses obtained from a given domain (Figure~\ref{fig:calib_figure2}). 

The cross-domain calibration curve shows that for NLU system without the calibration (baseline, R1 schemes), outputs scores that are over confident as the curve is above the diagonal line, particularly for scores close to zero. 
Given a rejection threshold, this would introduce unnecessary false rejects, since a sizable proportion of hypotheses are correct in those bins. 
After introducing the calibration objective (in R2, R3), our model’s reliability curve is nearly diagonal and a rejection threshold can be safely applied to such models. 
Note that the reliability curve does not reflect the accuracy (SemER) aspect of models as the number of hypotheses in each bin varies across different models. 
The per-domain calibration curves in Figure~\ref{fig:calib_figure2} for Global, HomeAutomation and Weather domains present a similar story where a domain's hypotheses tend to be over confident.
The calibration for baseline and R1 schemes is particularly off at low bins which may lead to a large number false rejects. 
Introduction of calibration in R2, R3 improve the calibration results, although the calibrated curves per domain are not as well aligned to the diagonal line as the cross-domain case.
This analysis suggests that our hypotheses generation strategy in section~\ref{sec:merge} may tend to break at lower thresholds as performance across domains are not calibrated. 
We also suggest that a different rejection threshold is applied to the top hypotheses, conditioned on the domain, since the calibration is different for each domain.

\begin{table}[t]
\centering
\begin{tabular}{ l|cc } 
 Optimization scheme & Relative improvements  \\ \hline
 Baseline & -            \\ 
 R1                   & 2.7\%             \\ 
 R2                   & 3.5\%             \\
 R3                   & 3.7\%            \\ 
\end{tabular}
\caption{Improvements in aggregated SemER after adding re-ranker scheme to the modularized NLU model. Baseline model's SemER is 7.1\%.}
\label{tab:results1}
\end{table}

\section{Additional experiment: Re-ranker training across domains with dissimilar data}
In all the previous experiments, we used the same set for training the re-rankers weights $\bm w_d$ across all domains. 
However, we can allow for further de-synchronization with an option of using different training/development data for different domains.
We simulate this by allowing a given domain to obtain a different subset of training/development portion from all of the available data.
Each domain samples $\sim$90\% of all the available training/development data.
Test partition is kept constant for cross-domain evaluation.
The re-ranker is trained using the R3 optimizations discussed in section~\ref{sec:rmo}.

Although we use different dataset per domain, we do not observe any significant degradation in performance of the NLU system (0.05\% relative to the re-ranker trained using R3 optimization on all of the development set).
This observation hints that further independence in curating training data for training each domain's recognizers.
We note that this asynchrony in training sets can only be allowed due to the calibrated nature of the re-ranker.
Although the initial direction is promising, we acknowledge the need to conduct more experiments as calibration may break under conditions involving very different re-ranker training datasets per domain. 


\section{Conclusion}
Large scale NLU models are often modularized for scalable and parallelized training. 
However, this requires a combination strategy, that can combine outputs across the domain specific components. 
We propose a modularized re-ranker model in this paper, trained to rank domain hypotheses based on their correctness, while yielding calibrated scores across domains.
The re-ranker design can allow asynchronous training per domain with implications towards a faster NLU model training and update. 
We demonstrate the usefulness of the re-ranker model on an Alexa use case, obtaining significant improvements in the SemER performance. 
We also present our findings on an experiment involving re-ranker trained on different datasets obtain by every domain independently. 
We observe that we are capable of maintaining the performance of the NLU system despite training models on different datasets, given the modularized and calibration aspects of the re-ranker.

In the future, we aim to add additional domain specific cues along with the IC, NER and DC scores (e.g. scores related to Named entity slots) to improve the generation of hypothesis list.
Similarly, more features (e.g. ASR based features \cite{aradilla2006using}, paralinguistic features \cite{gupta2014variable}) can be used to improve hypotheses re-ranking.
Given more features, we also aim to train a non-linear function $f$ to obtain the scores $s^{ui}_{\mathsf d}$.
Other optimization schemes for re-rankers can also be explored which combine existing loss functions and a calibration objective \cite{gavsic2015policy, kim2018scalable}.

\bibliographystyle{IEEEbib}
\bibliography{strings}

\end{document}